\documentclass[10pt,twocolumn,letterpaper]{article}

\usepackage{wacv}
\usepackage{times}
\usepackage{epsfig}
\usepackage{graphicx}
\usepackage{amsmath}
\usepackage{amssymb}
\usepackage{makecell}
\usepackage{booktabs}
\usepackage{url}      
\makeatletter
\g@addto@macro{\UrlBreaks}{\UrlOrds}
\makeatother
\usepackage{colortbl}

\newcommand{\Expect}{\mathbb{E}}
\newcommand{\KL}{\text{D}_\text{KL}}

\newcommand{\X}{\mathcal{X}}

\newcommand{\Y}{\mathcal{Y}}

\DeclareMathOperator*{\minimize}{min}

\usepackage[T1]{fontenc}
\usepackage[latin1]{inputenc}
\usepackage[table]{xcolor} 

\usepackage{float}

\newcommand{\brac}[1]{\left[#1\right]}

\wacvfinalcopy % *** Uncomment this line for the final submission

\def\wacvPaperID{17} % *** Enter the wacv Paper ID here

\ifwacvfinal\fi
\begin{document}

%Paper length is 8 pages for text + unlimited for references.  
%Double blind.  
%Submit publication approval request 5 working days prior to submission.  

%Due date is July 27.  

%%%%%%%%% TITLE
\title{SketchTransfer: A Challenging New Task for Exploring Detail-Invariance and the Abstractions Learned by Deep Networks}

% Authors at the same institution
%\author{First Author \hspace{2cm} Second Author \\
% Authors at different institutions
\author{Alex Lamb \\
MILA\\
{\tt\small lambalex@iro.umontreal.ca}
\and
Sherjil Ozair \\
MILA\\
{\tt\small sherjilozair@gmail.com}
\and
Vikas Verma \\
MILA, Aalto University, Finland\\
{\tt\small vikas.verma@aalto.fi}
\and
David Ha \\
Google Brain\\
{\tt\small hadavid@google.com}
}

\maketitle
\ifwacvfinal\thispagestyle{empty}\fi

%%%%%%%%% ABSTRACT
\begin{abstract}
Deep networks have achieved excellent results in perceptual tasks, yet their ability to generalize to variations not seen during training has come under increasing scrutiny.  In this work we focus on their ability to have invariance towards the presence or absence of details.  For example, humans are able to watch cartoons, which are missing many visual details, without being explicitly trained to do so.  As another example, 3D rendering software is a relatively recent development, yet people are able to understand such rendered scenes even though they are missing details (consider a film like Toy Story).  The failure of machine learning algorithms to do this indicates a significant gap in generalization between human abilities and the abilities of deep networks.  We propose a dataset that will make it easier to study the  detail-invariance problem concretely.  We produce a concrete task for this: SketchTransfer, and we show that state-of-the-art domain transfer algorithms still struggle with this task.  The state-of-the-art technique which achieves over 95\% on MNIST $\xrightarrow{}$ SVHN transfer only achieves 59\% accuracy on the SketchTransfer task, which is much better than random (11\% accuracy) but falls short of the 87\% accuracy of a classifier trained directly on labeled sketches.  This indicates that this task is approachable with today's best methods but has substantial room for improvement.  
\end{abstract}

\begin{figure}[htp!]
    \centering
    {\large Plane Class}
    \includegraphics[width=\linewidth,trim={0 15.0cm 0cm 0cm},clip]{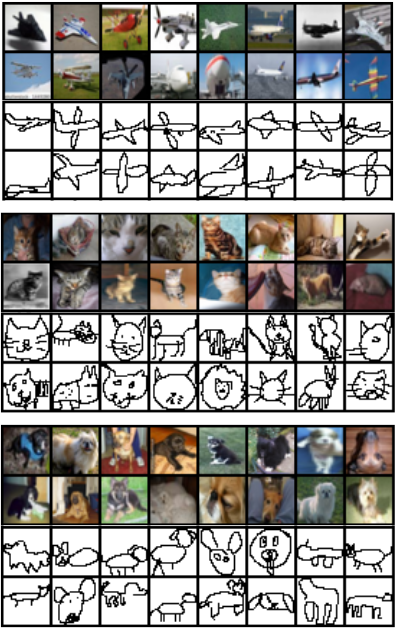}
    {\large Cat Class}
    \includegraphics[width=\linewidth,trim={0 7.5cm 0cm 7.4cm},clip]{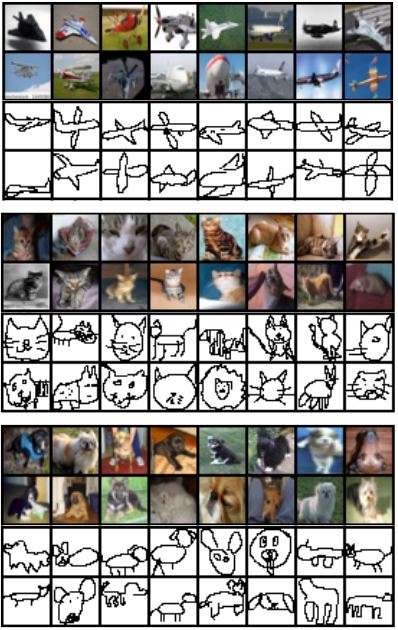}
    {\large Dog Class}
    \includegraphics[width=\linewidth,trim={0 0cm 0cm 15cm},clip]{figures/examples_sketchtransfer.png}
    \caption{The SketchTransfer task involves recognizing sketches while only having access to labeled examples of real images and unlabeled sketch images.  This is a challenging task for today's state-of-the-art transfer learning algorithms.  }
    \label{fig:examples_sketchtransfer}
\vspace{0.0cm}
\end{figure}

%%%%%%%%% BODY TEXT
\section{Introduction}

%Intro argument: 
%  -Humans learn abstractions that focus only on relevant aspects and are invariant to details.  
%  -Neural nets seem to have the opposite characteristic.  Adversarial examples, fooling examples, frequency.  
%  -We create a dataset where a neural network is evaluated on the quality of its abstractions learned without explicit supervision. 
%  -While ideally a network could perform this task with just real data, this may make the task too difficult.  
%  -To relax the task, we give the network access to unlabeled sketches with the goal of creating networks that can perform well with as little sketch data as possible (in practice we found the task is challenging even with a massive amount of sketch data).  
%  -

%Need to properly source this image if we do use it.  
%maybe even get permission.  
%https://aeon.co/essays/your-brain-does-not-process-information-and-it-is-not-a-computer
%todo: flip the sides.  
\begin{figure*}[t!]
    \centering
    \includegraphics[width=\linewidth]{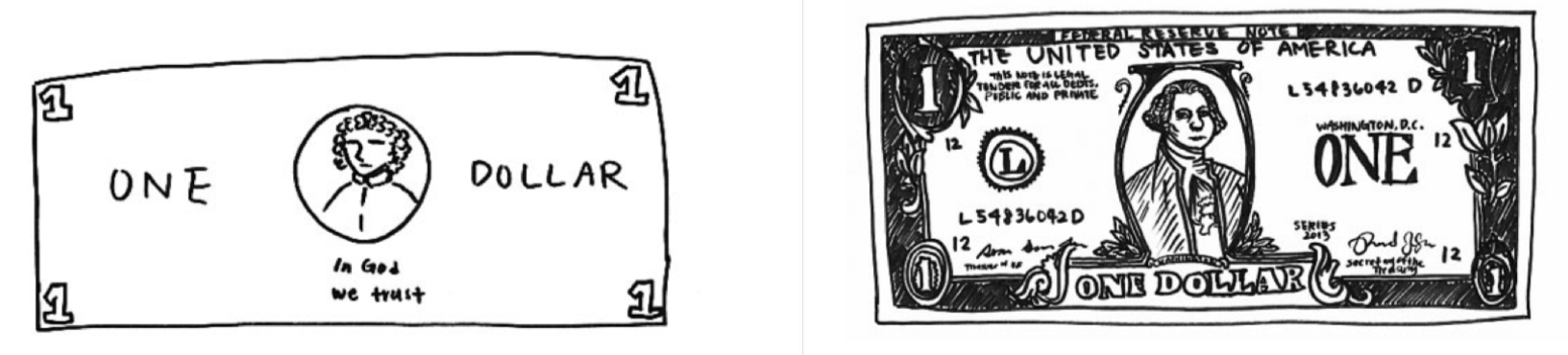}
\vspace{0.2cm}
    \caption{On the left, a person's sketch of a dollar bill from memory.  On the right, the same person's sketch with access to a reference.  This indicates that humans only remember and can perform recognition with only a small handful of salient aspects of data and have substantial detail invariance.  (Image credit: \textit{The empty brain}~\cite{epstein2016empty})}
    \label{fig:dollarsketch}
\end{figure*}

%AG: May be directly give an example of cartoons or sketches in the openning paragraph ?

%AG: I think, introduction needs to be improved. Why you are actually creating SketchTransfer dataset is still not that clear. You are not trying to solve every problem as you mentioned in the introduction, so ideally some details should be removed from introduction and moved to related work.

%AG: May be interesting thing to note is, what problem you are actually solving in the paper. Its not clear in the first 6 paragraphs.

%AG: May be one angle to use, is to say, that we want to create a dataset using which we can "transfer" different amount of details, and test based on that. Its not straightforward how one can extend this for MNIST/CIFAR/SVHN. People do this paradigm for RL, but thats the other extreme end.
Humans experience a high resolution world which is rich in details, yet are able to discern small amounts of highly relevant information from the world.  Thus humans learn abstractions that store important pieces of information while discarding others.  For example, Figure~\ref{fig:dollarsketch} shows how only a subset of the visual details in a dollar bill are remember by a person asked to produce a sketch without a reference.

%AG: It's an interesting comment. You can cite something here.
The quality of the abstractions learned by humans may be an important factor in the flexibility and adaptability of human perception: we are able to understand objects and our environment even as details change.  

%AG: Cite more stuff.
Disturbingly, a growing body of evidence shows that while deep networks have visual perception which is competitive with humans (at least in some cases), the abstractions learned by deep networks differ substantially from what humans learn. 

An exploration by \cite{wang2019high} showed that deep networks correctly classify objects when only high-frequency components of the image are preserved but fail completely when only low-frequency components are preserved (the low frequency version of image is a blurred version of that image).  This is the opposite of the inductive bias demonstrated by humans.  

The adversarial examples literature has \cite{szegedy2013intriguing} found that neural network's predictions can be changed by extremely small (adversarially selected) perturbations which are imperceptible to human vision.  At the same time these small perturbations have either a very small effect on humans (when given very little time to process an image) \cite{elsayed2018human}  or no effect at all.  

The BagNet project \cite{brendel2018bagnet} explored using bags of local features for Imagenet classification and found that such models could achieve  competitive results.  Additionally, they found that if one performed style transfer on real images \cite{gatys2015neural} and retained the textures while scrambling the image's content, standard convolutional imagenet classifiers retained strong performance.  This is evidence that convnets trained on datasets similar to Imagenet may learn primarily from local textures while discarding global structure, indicating a strong lack of detail invariance.  
On artificial images with the local texture from one image but the global shape and content of another image, imagenet classifier's nearly always prefer the class associated with the texture rather than the content \cite{geirhos2018imagenet}.  Additionally this can be partially addressed and generalization improved by training on artificially stylized examples in the imagenet dataset.  

We create a dataset, which we call SketchTransfer, where a neural network is evaluated on the quality of its abstractions learned without explicit supervision.   More specifically, we constructed a dataset in which a model must be able to classify images of sketches while only having access to labels for real images.  As an example of why this task is difficult, consider sketches of dogs and cats.  Many of these sketches choose to focus on the face of the animal.  Oftentimes the only clear difference between a dog sketch and a cat sketch is the shape of the ears, with dogs having round ears and cats having pointed ears.  While the shape of the ears is a feature which is present in real images of cats and dogs, a neural network may not pick up on this highly salient feature.  

While ideally a network could perform this task with just real data and without any access to data points with missing details, this may make the task too difficult.  To make the task easier, we give the network access to unlabeled sketches with the long-term goal of creating networks that can perform well with as little sketch data as possible or perhaps no sketch data at all (in practice we found the task is challenging even with a substantial amount of sketch data).  

An emerging view in machine learning is that a great deal of learning is accomplished in a ``self-supervised way'' by relying primarily on the structure in unlabeled natural data as a form of supervision.  A successful algorithm on the SketchTransfer dataset could provide evidence that detail invariance is also achievable without the use of explicit of supervision of data with missing details.  

%(although the use of actual sketches in the unlabeled dataset is optimistic). 

Our new dataset, which we call SketchTransfer, provides the following contributions: 

\begin{itemize}
    \item A new dataset for transfer learning, which is approachable with today's state of the art methods but is still very challenging.  
    \item A demonstration that even well-regularized algorithms fail to generalize across different levels of detail.  
    \item A brief exposition of a few state-of-the-art approaches to transfer learning and experiments with them on the SketchTransfer dataset.  
\end{itemize}

\section{SketchTransfer}

%Class Correspondence.  
%Examples of classes from each.  

The SketchTransfer training dataset consists of two parts: labeled real images and unlabeled sketch images.  The test dataset consists of labeled sketch images.  To make the task as straightforward as possible, we used the already popular CIFAR-10 dataset as the source of labeled real images.  

We used the quickdraw dataset \cite{quickdraw2019dataset} as the source of sketch images.  This dataset consists of 345 classes and was collected by asking volunteers to quickly sketch a given class with a 20 second time limit.  For the SketchTransfer dataset, we rendered the QuickDraw images at a fixed resolution of 32x32.  

The quickdraw dataset contains many more classes than CIFAR-10 and the classes are slightly different and generally more detailed.  To solve this we defined a correspondence between the classes in CIFAR-10 and a relevant subset of the quickdraw classes (shown in Table~\ref{tb:correspondence}).  For example we map the quickdraw classes ``Car'' and ``Police Car'' onto the CIFAR-10 class automobile.  

One small issue is that CIFAR-10 contains a deer class and QuickDraw doesn't.  Thus we elected to make our SketchTransfer test set only have 9 classes, but for the convenience of researchers we keep the CIFAR-10 training set the same (having 10 classes).  Our final dataset of sketches consists of 9 classes which correspond directly to CIFAR-10 classes.  This sketch dataset has a total of 90000 training images and 22500 test images (10000 and 2500 per class respectively).  

\begin{figure}
    \centering
    Plane
    \includegraphics[width=\linewidth,trim={0 0 12cm 0},clip]{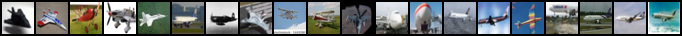}
    \includegraphics[width=\linewidth,trim={0 0 12cm 0},clip]{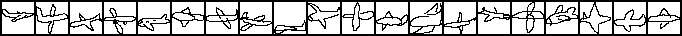}
    Car
    \includegraphics[width=\linewidth,trim={0 0 12cm 0},clip]{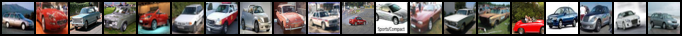}
    \includegraphics[width=\linewidth,trim={0 0 12cm 0},clip]{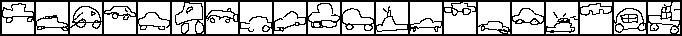}
    Bird
    \includegraphics[width=\linewidth,trim={0 0 12cm 0},clip]{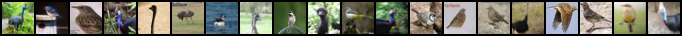}
    \includegraphics[width=\linewidth,trim={0 0 12cm 0},clip]{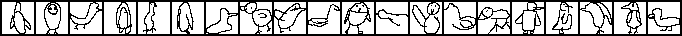}
    Cat
    \includegraphics[width=\linewidth,trim={0 0 12cm 0},clip]{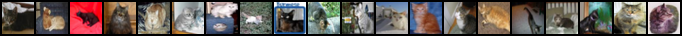}
    \includegraphics[width=\linewidth,trim={0 0 12cm 0},clip]{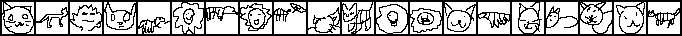}
    Dog
    \includegraphics[width=\linewidth,trim={0 0 12cm 0},clip]{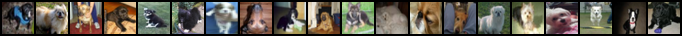}
    \includegraphics[width=\linewidth,trim={0 0 12cm 0},clip]{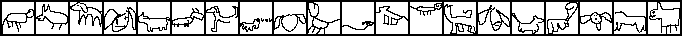}
    Frog
    \includegraphics[width=\linewidth,trim={0 0 12cm 0},clip]{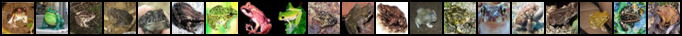}
    \includegraphics[width=\linewidth,trim={0 0 12cm 0},clip]{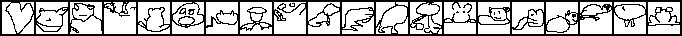}
    Horse
    \includegraphics[width=\linewidth,trim={0 0 12cm 0},clip]{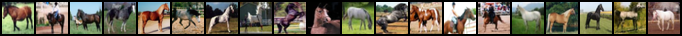}
    \includegraphics[width=\linewidth,trim={0 0 12cm 0},clip]{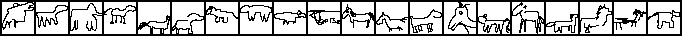}
    Ship
    \includegraphics[width=\linewidth,trim={0 0 12cm 0},clip]{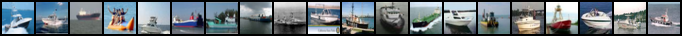}
    \includegraphics[width=\linewidth,trim={0 0 12cm 0},clip]{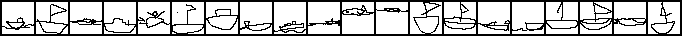}
    Truck
    \includegraphics[width=\linewidth,trim={0 0 12cm 0},clip]{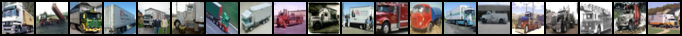}
    \includegraphics[width=\linewidth,trim={0 0 12cm 0},clip]{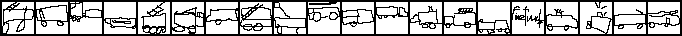}
    \caption{Examples from all of the classes in the SketchTransfer dataset, with real images from the class and sketch images from the class.  }
    \label{fig:examples_of_classes_in_dataset}
\end{figure}

Examples from both real and sketch images across all the classes can be seen in Figure~\ref{fig:examples_sketchtransfer} and Figure~\ref{fig:examples_of_classes_in_dataset}.  One important property of SketchTransfer is that aside from sharing common classes, there is no explicit pairing between the real images and the sketch images.  

We received permission from the Google Creative Lab to use the Quickdraw dataset \cite{quickdraw2019dataset}.   

\section{Related Datasets}

Datasets of human sketches make it possible for algorithms to learn representations closely aligned with human priors. Prior to QuickDraw~\cite{quickdraw2019dataset}, the Sketch dataset~\cite{berlindb} of 20K hand sketches was used to explore feature extraction techniques.  A later work, the Sketchy dataset~\cite{sketchydb}, provided 70K vector sketches paired with corresponding photo images for various classes to facilitate a larger-scale exploration of human sketches. ShadowDraw~\cite{shadowdraw} used a dataset of 30K raster images combined with extracted vectorized features to construct an interactive system that predicts what a finished drawing looks like based on a set of incomplete brush strokes from the user's digital canvas. In addition to human sketches, ancient Eastern scripts also have sketch-like properties. For example, \cite{clanuwat2018deep} considered using an RNN and VAEs to generate sketched Japanese characters.  Understanding and recognizing these historical Japanese characters has become an important and widely studied problem in the digital humanities field \cite{clanuwat2018kuzu,clanuwat2019kuronet}. 

In developmental childhood psychology, \cite{long2018drawings,long_fan_chai_frank_2019} performed a series of experiments in which they tasked children of various ages with drawing sketches of certain classes, and then studying the properties of the sketches using convolutional neural networks.  Their main finding was that visual features become more distinctive in the drawings produced by older children and that this goes beyond differences in visuomotor controls.  This provides some evidence that for humans the deeper knowledge gained about objects from growing older, is reflected in sketches.

{\renewcommand{\arraystretch}{1.3}
\begin{table}[htp!]
\centering
\caption{Class correspondences used to construct the sketch part of the SketchTransfer dataset.  }
\label{tb:correspondence}
\vspace{0.2cm}
\rowcolors{2}{gray!25}{white}
\begin{tabular}{ccc}
\rowcolor{gray!50}
Cifar-10 Class & QuickDraw Classes\\
\makecell[c]{Airplane} & \makecell[c]{Airplane} \\
Automobile & \makecell[c]{ Car\\\quad\quad\quad\,\, Police Car \quad\quad\quad\quad} \\
\makecell[c]{Bird} & \makecell[c]{Bird\\Duck\\Flamingo\\Owl\\Parrot\\Penguin\\Swan} \\
 Cat & \makecell[c]{Cat\\Lion\\\quad\quad\quad\quad\,\, Tiger \,\quad\quad\quad\quad\quad}\\
Deer & \makecell[c]{n/a}\\
Dog & Dog \\
\makecell[c]{Frog} & \makecell[c]{Frog}\\
Horse & Horse\\
\makecell[c]{Ship} & \makecell[c]{Cruise Ship\\Sailboat\\Speedboat}\\
Truck & \makecell[c]{\quad\quad\quad\quad\,\, Truck \,\,\,\,\,\,\quad\quad\quad\quad\\ \quad  Firetruck \,\,}\\
\end{tabular}
\vspace{-0.5cm}
\end{table}}

\section{Baselines}
\label{sec:baselines}

We evaluated several techniques on the proposed SketchTransfer dataset.  First we considered techniques which only use the labeled source data (real images) including the use of regularization.  As these models don't see the sketches during training, we might expect that it would be difficult for these models to generalize well on sketches.  For techniques which take advantage of the unlabeled data, we primarily focus on methods which perform well on existing benchmarks.  One such benchmark is SVHN to MNIST, which involves transfer learning between very detailed color images of digits (SVHN) and plain black-and-white digit images (MNIST) \cite{mao2019virtual}.  One key idea is to try to make the model's representations on source datapoints and target domain datapoints more similar, which we discuss in Section~\ref{sec:adapt}.  Another key idea is to enforce a consistency based objective on the target domain, which is also widely used in semi-supervised learning \cite{verma2019ict}.  The idea with this is to discourage the classifier's decision boundary from passing through regions where the data in the target domain has high density (and thus encourage connected regions of high density to share the same predicted label).  Virtual adversarial training (Section~\ref{sec:vat}) and virtual mixup training (Section~\ref{sec:vmt}) are both motivated by this notion of consistency.  

\subsection{Training only on Labeled Source Data}

As the simplest baseline, we consider simply training a network on the CIFAR-10 dataset and evaluating on sketches.  

\subsection{Regularized Training on Source Data}

One possible approach is to still only use the real labeled images, but use regularization.  A possibility is that this will force the model to use the more salient features in the data to predict on real images which will then be usable when predicting on sketches.  For example, \cite{verma2019manifold} found that the use of Manifold Mixup improves robustness to several classes of artificial image distortions.  

\begin{align}
\begin{split}
\label{eq:mixup}
  \tilde{x} &= \lambda x_i + (1 - \lambda) x_j,\\
  \tilde{y} &= \lambda y_i + (1 - \lambda) y_j.
\end{split}
\end{align}

Mixup is a state of the art regularizer for deep networks which involves training with linear interpolations of images and using a corresponding interpolation of the labels (Equation~\ref{eq:mixup}).  In its original formulation \cite{zhang2017mixup}, it is only applicable to training with labeled data, although an unsupervised variant has also been explored \cite{beckham2019adversarial}.  

% DH: equation above not used in the same section, maybe some more explanation here?

\subsection{Adversarial Domain Adaptation}
\label{sec:adapt}

Domain adversarial training \cite{ganin2016domain} consists of taking the hidden representations learned by a network on a source domain and a target domain and using an adversarially trained discriminator $d$ to make the representations $g(x)$ follow the same overall distribution.  Since the discriminator is trained on hidden states (for both source and target domains), it does not require labeled data from the target domain.  

\begin{multline}
\label{eq:dann}
\minimize_{f}  \mathcal{L}_y(f;\X_s, \Y_s) + \lambda_d \mathcal{L}_d(g;\X_s,\X_t)\\
\mathcal{L}_d(g;\X_s,\X_t) = \sup_{d}\Expect_{x \sim \X_s} \brac{\ln d(g(x))} \\+ \Expect_{x \sim \X_t} \brac{\ln (1 - d(g(x)))},
\end{multline}

The adversarial domain adaptation objective (Equation~\ref{eq:dann}) augments the usual classifier objective on source data $\X_s$ and a hyperparameter $\lambda_d$ is used to adjust the weight of the adversarial loss, which aims to make features $g(x)$ look similar for examples from the source and target distributions.  

\subsection{Virtual Adversarial Training}
\label{sec:vat}

The virtual adversarial training (VAT) \cite{miyato2018virtual} algorithm is based on the Consistency regularization principle.  The central idea of this algorithm is to use pseudolabels of unlabeled samples to generate adversarial perturbations, and use these adversarial perturbations instead of random perturbations for Consistency regularization.  
We always used the FGSM attack for computing the virtual adversarial perturbations \cite{goodfellow2014explaining}.  

\begin{align}
\mathcal{L}_v(f; \X) = \Expect_{x \sim \X} \brac{\max_{\| r\| \le \epsilon} \KL(f(x) \| f(x + r))}.
\label{eq:vat}
\end{align}

Thus given some ball with a radius $\| r\| \le \epsilon$ around each of the data points we encourage the change in the model's prediction $f$ to be small (Equation~\ref{eq:vat}).  

%The use of virtual adversarial training along with adversarial domain adaptation (Section~\ref{sec:adapt}) was introduced as ``Virtual Adversarial Domain Adaptation'' (VADA) \cite{shu2018dirt}.  

\subsection{Dirt-T}

The Decision Boundary Iterative Refinement Training with a Teacher (or ``Dirt-T'') algorithm \cite{shu2018dirt} starts with a pre-trained model and considers an iterative refinement procedure to enforce the cluster assumption on the unlabeled target data (the cluster assumption is the notion that decision boundaries are more likely to pass through regions of low-density).  Because this is only done on the target domain, it leads to high loss on the source domain, but also allows for the possibility of distribution-shift between the source and target domains.  

\subsection{Virtual Mixup Training (VMT)}
\label{sec:vmt}

The approaches which combine \textit{consistency regularization} with Mixup training \cite{mixup, verma2019manifold,verma2019graphmix} have shown to achieve state-of-the-art results in semi-supervised learning paradigm \cite{verma2019ict,mixmatch}. VMT \cite{mao2019virtual} extends these approaches in the paradigm of Unsupervised Domain Adaptation by combining them with the Domain Adversarial Training \cite{ganin2016domain}. More formally, VMT \cite{mao2019virtual} augments the loss function of Domain Adversarial Training and the entropy minimization loss with the Mixup loss.

In the case of target domain, since the targets are not available, similar to the \cite{verma2019ict, mixmatch, miyato2018virtual}, the pseudolabels of the unlabeled samples are used for mixing. The mixing of samples and pseudolabels are done as follows (where $\lambda~\sim Beta(\alpha, \alpha)$):
\begin{align}
\label{eq:vmt}
\begin{split}
  \tilde{x} &= \lambda x_i + (1 - \lambda) x_j,\\
  \tilde{y} &= \lambda f(x_i) + (1 - \lambda) f(x_j),
\end{split}
\end{align}

The Mixup loss function on pseudolabels (Equation~\ref{eq:vmt}), that can be augmented with other losses, is  given as follows:
\begin{align}
\label{eq:vmt_loss}
\mathcal{L}_m(f; \X) = \Expect_{x \sim \X} \brac{ \KL(\tilde{y} \| f(\tilde{x}))}.
\end{align}

%This overall objective is shown in Equation~\ref{eq:vmt_loss}.  

\subsection{Rotation Prediction}

We can also consider adding self-supervised objectives, with the intuition that solving this objective may require the model to have a more thorough understanding of salient aspects than is required to predict the class.  One such self-supervised objective that we consider, due to its simplicity, is randomly rotating input images by either 0 degrees, 90 degrees, 180 degrees, or 270 degrees, and having the model classify the degree of rotation as a 4-way classification task \cite{feng2019rotate}.  As this objective does not require labels, we can apply it on both the real images and the sketch images.  

\subsection{CyCada and CycleGAN}

The CyCada project \cite{hoffman2017cycada} explored the use of the CycleGAN algorithm \cite{zhu2017unpaired} for transfer learning to new domains.  Essentially the CycleGAN consists of an encoder which maps from the source domain to the target domain and a decoder which maps back to the source domain.  A discriminator is used to encourage the encoded domain to follow the same distribution as the target domain.  Notably this can be done without having access to paired examples from the source and target domains.  

%\subsection{Invariant Risk Minimization (IRM)}

%\subsection{Training on restyled CIFAR-10 images}

%\cite{geirhos2018imagenet}.  

%\subsection{IRM + Self-supervised learning}

%Idea: do IRM but on a self-supervised objective, since the self-supervised objective will have targets on both the source and the target domain.  

%The basic intuition is that if our features let us predict different parts of the structure (for example, given the left side of a person, predicting what the right side will look like) and does so on both the sketch and real images, then it will learn good features for the sketches.  

%\subsection{Training after Edge Detection}

%An important misconception is that sketches are similar to edges from realistic images.  

\section{Experiments}

First we consider running all the baselines discussed in Section~\ref{sec:baselines} on the full SketchTransfer dataset, with 10000 unlabeled images per class (or zero per class, for the baselines which don't use the unlabeled sketches).  All numbers are accuracies.  We ran two identical trials of each experiment with different random seeds and report the mean and standard deviation.  The results of running with these baseline algorithms is presented in Table~\ref{tb:results1}.  

We note that our best result not using labeled sketch data is the same combination of techniques which achieved the best result in Virtual Mixup Training \cite{mao2019virtual}.  However on their tasks involving MNIST to SVHN and SVHN to MNIST transfer, this approach was able to almost match the performance of a classifier trained on labeled examples from the target domain.  However on the SketchTransfer task there is still a clear gap between the 59\% obtained with only unlabeled sketches and the 87\% obtained by training on labeled sketches.  

%Method, train acc, test acc.  
%\begin{table*}[htp!]
%\centering
%\caption{Performance of different baseline models on SketchTransfer.  We ran multiple trials of each and report mean and standard deviation.  }
%\vspace{0.2cm}
%\label{tb:results0}
%\begin{tabular}{lrrr} %\hline
%\toprule
%Method & \makecell{Unlabeled Sketches\\ Per Class} & Model & Test Accuracy \\ %\hline
%\midrule
%VMT+DIRT-T (40 dirt-t epochs) & 10000 & cnn-small & 61.7,56.0 \\
%VMT-no-DIRT-T & 10000 & cnn-small & 45.3, 57.9 \\
%VMT-noDirtT only domain adv & 10000 & cnn-small & 30.13 \\
%VMT-noDirtT just v-mixup & 10000 & cnn-small & 22.5 \\
%VMT-noDirtT just VAT & 10000 & cnn-small & 40.7 \\ %run21
%Source Mixup and VAT (no use of target data) & 0 & 22.8 & 23.6 \\ %note need to run longer
%VMT-no-DirtT + predict rotation & 10000 & small & & \\
%CyCada & 10000 & ResNet & 25.15 \\
%IRM & 10000 & cnn-small & \\
%Source no instance normalization & 0 & cnn-small-nobn & 22.92, 23.66 \\
%Source no instance normalization self-supervised rot-pred & 10000 &  cnn-small-nobn & \\
%Source no mixup no vat & 0 & cnn-small & 37.4 \\
%Source w/ mixup no vat & 0 & cnn-small & 35.5 \\
%Source no mixup w/vat & 0 & cnn-small & 36.3 \\
%Source w/ mixup w/vat & 0 & cnn-small & 37.96 \\
%Train on Quickdraw Supervised & n/a & cnn-small & 86.8 \\ %run18
%\bottomrule
%\end{tabular}
%\end{table*}

%\setlength{\tabcolsep}{20pt}
\renewcommand{\arraystretch}{1.5}{
\begin{table*}[]
\centering
\caption{Performance of different baseline models on SketchTransfer.  We ran two trials for each experiment and we report mean with standard deviation in parenthesis.  We also report a supervised learning result for a model trained directly on the labeled sketch training set.  Our results indicate that using the unlabeled sketch data improves substantially but still falls short of the a model trained directly on labeled sketches.  }
\vspace{0.2cm}
\label{tb:results1}
\rowcolors{2}{gray!25}{white}
\begin{tabular}{lcccccccccc} %\hline
\toprule
\rowcolor{gray!50}
\makecell{Inst.\\Norm} & \makecell{Domain\\Adv.} & \makecell{Source\\Mixup} & \makecell{Virtual\\Mixup} & \makecell{Source\\VAT} & \makecell{Target\\VAT} & DIRT-T & CyCada & \makecell{Rotation\\Pred.} & \makecell[c]{\, Unlabeled\\ \, Sketches\\ \, Per Class} & \makecell[c]{\quad\quad Test\quad\quad\quad\\Accuracy} \\ %\hline
%\midrule
\checkmark  &   &   & \checkmark &   &   &   &   &   & 10000 & $17.55 (4.95)$ \\%\arrayrulecolor{gray}%\hline
 %12.6, 22.5\\ \hline
   &   &   &   &   &   &   &   &   &   0   & $23.29  (0.37)$ \\%\arrayrulecolor{gray}%\hline % 22.92, 23.66\\
   &   &   &   &   &   &   & \checkmark &   & 10000 & $25.27 (0.12)$\\%\arrayrulecolor{gray}%\hline
   &   &   &   &   &   &   &   & \checkmark & 10000 & $27.90 (0.38)$ \\%\arrayrulecolor{gray}\hline %27.52,28.28 \\
\checkmark  & \checkmark &   &   &   &   &   &   &   & 10000 & $30.31 (0.19)$ \\%\arrayrulecolor{gray}\hline %30.13, 30.50\\
\checkmark  &   & \checkmark &   &   &   &   &   &   &   0   & $35.66 (0.16)$ \\%\arrayrulecolor{gray}\hline %35.50, 35.82\\
\checkmark  &   &   &   &   &   &   &   &   &   0   & $35.84 (1.56)$ \\%\arrayrulecolor{gray}\hline %37.40,34.28 \\
\checkmark  &   &   &   & \checkmark &   &   &   &   &   0   & $37.74 (1.44)$ \\%\arrayrulecolor{gray}\hline %36.30,39.17 \\
\checkmark  &   & \checkmark &   & \checkmark &   &   &   &   &   0   & $38.72 (0.75)$ \\%\arrayrulecolor{gray}\hline %37.96,39.47 \\
\checkmark  &   &   &   & & \checkmark   &   &   &   & 10000 & $41.15 (0.45)$ \\%\arrayrulecolor{gray}\hline %40.7, 41.60\\
\checkmark  & \checkmark & \checkmark & \checkmark & \checkmark & \checkmark &   &   &   & 10000 & $51.60 (6.30)$ \\%\arrayrulecolor{gray}\hline % 45.3,57.9 \\
\checkmark  & \checkmark & \checkmark & \checkmark & \checkmark & \checkmark & \checkmark &   &   & 10000 & $\underline{58.85 (2.85)}$ \\ %\textbf{61.7,56.0} \\
%\arrayrulecolor{black}\midrule
\checkmark  &   & \checkmark &   & \checkmark &   &   &   &   & Supervised & $87.25 (0.45)$ \\ %86.8, 87.7 \\  
\bottomrule
\end{tabular}
\vspace{-0.5cm}
\end{table*}}

%\subsection{Number of Unlabeled Sketches Required}
%What if we train with a greatly reduced number of sketches?  

%\subsection{Sketch2Image}

%While it is not the focus of this work, we briefly note that it's possible to consider the opposite task: training on labeled sketch images and unlabeled real images to learn a classifier that performs well on real images.  This task is perhaps more practically relevant.  

\subsection{Analysis}

%When model makes mistakes, which classes is it most likely to get confused

While we demonstrated in Table~\ref{tb:results1} that a model trained only on real images performs better than average on sketches, and this performance is improved by the use of transfer learning with sketches, it gives little direct insight into the nature of this improvement.  To address this we considered two different analysis on the learned models.  To make this analysis as clean as possible, we only considered our best model trained only on real images, our best model trained on unlabeled sketches, and our best model trained on label sketches.  The best model was selected according to the test accuracies in Table~\ref{tb:results1}.  All of these analysis were performed on the test set.  

\begin{figure}[!]
    \centering
    \includegraphics[width=1.0\linewidth]{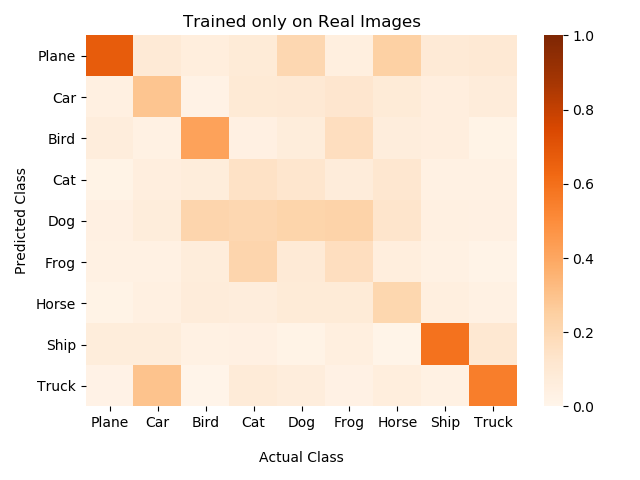}
    \includegraphics[width=1.0\linewidth]{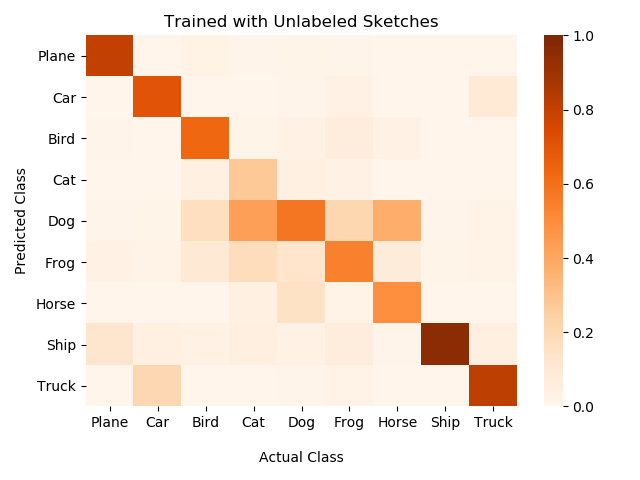}
    \includegraphics[width=1.0\linewidth]{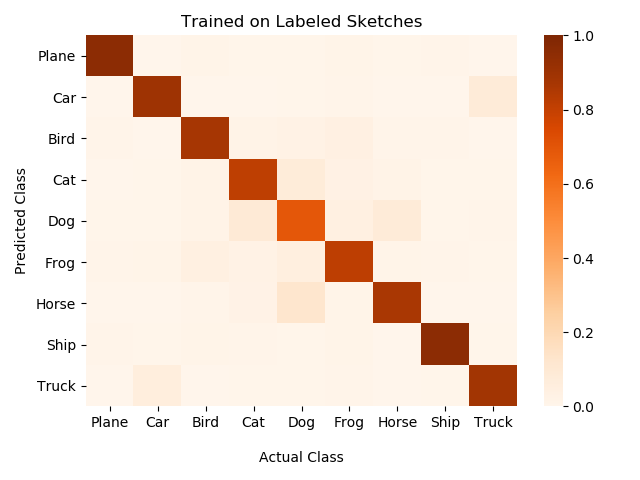}
    \caption{Confusion matrices indicating predicted and true classes for a model trained only on real images (top), our best transfer learning model (center), and a model trained on labeled sketch images (bottom).  A stronger diagonal indicates better performance.  The model trained only on real images has the best performance on planes and trucks.  The model exploiting unlabeled images performs better on many more classes, but still struggles to separate cats and dogs as well as horses and dogs.  }
    \label{fig:confusion}
\end{figure}

First we considered a confusion matrix analysis, given in Figure~\ref{fig:confusion}.  The confusion matrix shows the distribution over predicted classes for each ground truth class.  A stronger diagonal indicates higher accuracy.  We can see that a model trained only on real images is most accurate on planes, trucks, and ships.  The confusion matrix for the model trained with unlabeled sketches shows dramatically better classification on the bird, cat, dog, frog, and horse classes.  

\begin{figure}[ht!]
    \centering
    Pixel-Based
    \includegraphics[width=\linewidth,trim={0 18cm 0cm 1.2cm},clip]{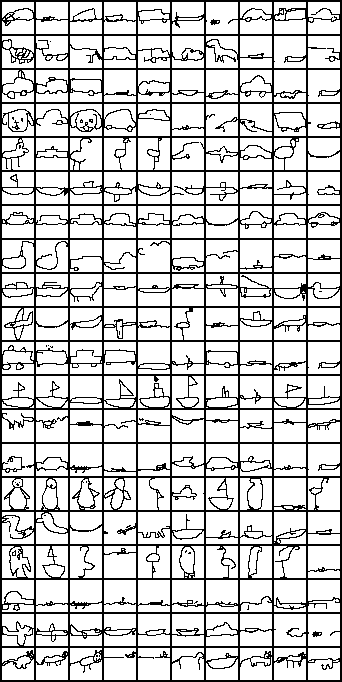}
    Randomly Initialized Network
    \includegraphics[width=\linewidth,trim={0 18cm 0cm 1.2cm},clip]{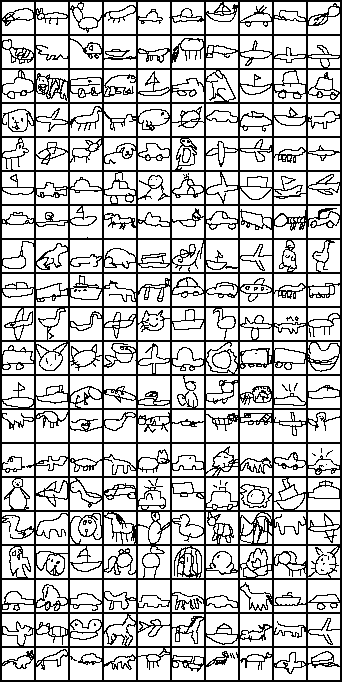}
    Network Trained only on Real Images
    \includegraphics[width=\linewidth,trim={0 18cm 0cm 1.2cm},clip]{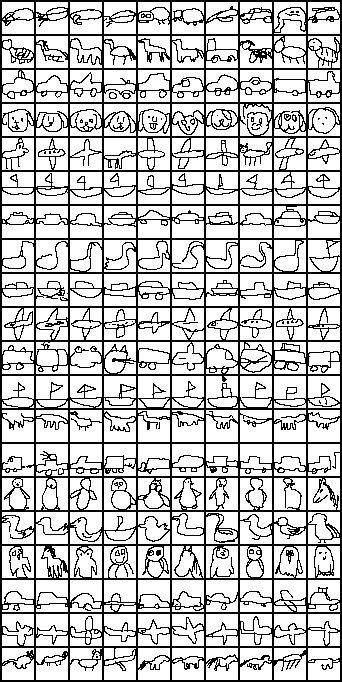}
    Network Trained with Unlabeled Sketch Images
    \includegraphics[width=\linewidth,trim={0 18cm 0cm 1.2cm},clip]{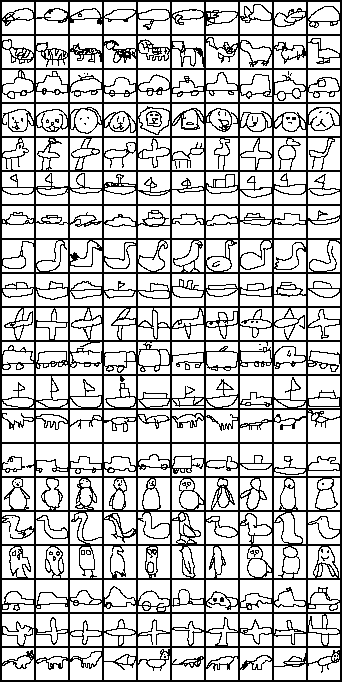}
    Network trained on Labeled Sketch images
    \includegraphics[width=\linewidth,trim={0 18cm 0cm 1.2cm},clip]{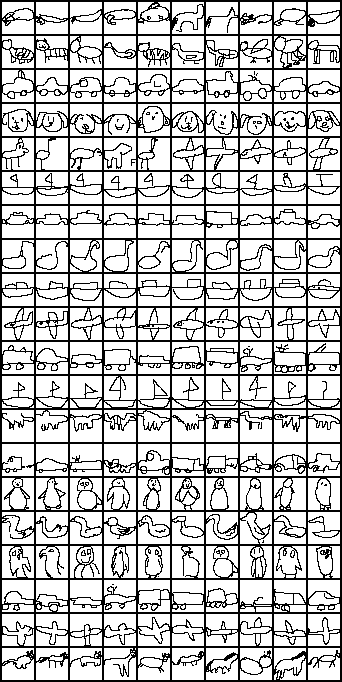}
    \caption{Nearest Neighbors on the test set using different distances (original on the left).  Note that the distance learned after transfer learning is clearly more semantically meaningful, and even a well-regularized model using only the source data learned surprisingly good features for sketches.  }
    \label{fig:nearest_neighbors}
\end{figure}

Additionally, we conducted an analysis where we collected the k-nearest neighbors for test images using different distance metrics (Figure~\ref{fig:nearest_neighbors}).  First, we considered a simple euclidean distance in pixel space, and this yielded extremely poor nearest-neighbors, with very low semantic and class similarity.  Next we used different encoders and ran it until reaching an 8x8 spatial representation and used simple euclidean distance in this space.  With a purely random encoder, this yielded poor results.  However, with networks trained on real images, the quality of the nearest neighbors was substantially improved in terms of semantic similarity.  We also observed that the network trained using transfer learning on unlabeled sketches had even more semantically relevant nearest neighbors.

\section{Discussion}

In this work, we propose the problem of whether a machine learning algorithm can develop representations of everyday things that can be mapped to human doodles, but giving the learning algorithm access to unlabelled doodle data. We could also consider an alternative approach where we train an artificial agent to sequentially draw, one stroke at a time, a given pixel image (such as CIFAR-10 samples) and by constraining the number of brush strokes our agent is allowed to make, we may be able to similar the biological constraints of the human anatomy and thus also develop a human doodle-like representation of actual images.

This approach has been explored in \textit{Artist Agent}~\cite{artistagent} and SPIRAL~\cite{ganin2018synthesizing} where an agent is trained to paint an image from a data distribution using a paint simulator. Subsequent works~\cite{zheng2018strokenet,huang2019learning} combined the SPIRAL algorithm with the sketch-rnn~\cite{ha2017neural} algorithm and enhanced the drawing agent by giving it an internal \textit{world model}~\cite{ha2018worldmodels} to demonstrate that the agent can learn to paint inside its imagination. While the aforementioned works train an agent with an objective function to produce doodles that are as photorealistic as possible, a recent work~\cite{nakano2019neural} trains an agent to optimize instead for the \textit{content loss} defined by a neural style transfer algorithm~\cite{gatys2015neural}, opening up the exciting possibility of training agents that can produce truly abstract versions of photos.

However, the doodle representations of everyday objects developed by humans are not merely confined to our constrained biological ability to draw objects sequentially with our hands, stroke-by-stroke, but cultural influence is also at play. The location meta-data from the original QuickDraw dataset provided interesting examples of this phenomenon. For instance, it has been observed that most Americans draw circles counterclockwise, while most people in Japan draw them clockwise~\cite{qz2017}. Chairs drawn in different countries tend to have different orientations~\cite{jana2017}. Snowman in hotter countries consists of two snowballs, while for colder countries consists tend to have three~\cite{forma2017}. By giving our algorithm actual unlabelled doodle data produced by humans around the world, we give our machine learning algorithms a chance to learn the true representation of objects developed by all of humanity.

\section{Future Work}

The idea of SketchTransfer is quite general and we identify a few ways in which it could be extended: 

\begin{itemize}
    \item While CIFAR-10 does contain realistic images, it is limited in a few ways.  First, the images are rather small, being only 32x32.  Second, the number of labeled images is relatively small.  Third, they generally only contain the object of interest and lack context, both spatial and temporal.  Selecting a dataset which is stronger along any of these axis could be a useful improvement to the SketchTransfer task.  
    \item Sketches are to a large extent an extreme case of images which lack irrelevant details.  Various cartoons and shaded illustrations may be a reasonable middle-ground and serve as an easier analogue task to SketchTransfer.  
    \item We elected to make the task easier by providing access to unlabeled sketch images.  Another way to accomplish this might be to provide a very small number of labeled sketch images and use meta-learning to perform few-shot (or one-shot) classification of sketches.  The task classification without using any sketch data during training would be challenging, yet \cite{goyal2019recurrent} demonstrated strong generalization to changing environments by encourage independent mechanisms as an inductive bias of the architecture.  
\end{itemize}

\newpage
\section{Conclusion}

Human perceptual understanding shows a great deal of invariance to details and emphasis on salient features which today's deep networks lack.  We have introduced SketchTransfer, a new dataset for studying this phenomenon concretely and quantitatively.  We applied the current state-of-the-art algorithms on domain transfer to this task (along with ablations).  Intriguingly we found that they only achieve 60\% accuracy on the SketchTransfer task, which is between the 11\% accuracy of a random classifier but falls dramatically short of the 90\% classifier trained on the labeled sketch dataset.  This indicates that this new SketchTransfer task could be a powerful testbed for exploring detail invariance and abstractions learned by deep networks.

\clearpage
{\small
\bibliographystyle{ieee}
\bibliography{egbib}
}

\end{document}